\documentclass[runningheads]{llncs}
\usepackage{graphicx}

\usepackage{tikz}
\usepackage{comment}
\usepackage{amsmath,amssymb} 
\usepackage{color}

\usepackage{multirow}
\usepackage{xcolor}
\usepackage{adjustbox}
\usepackage[font=small,labelfont=bf]{caption}
\usepackage[accsupp]{axessibility}  

\usepackage[toc,page]{appendix}
\usepackage[linesnumbered,ruled]{algorithm2e}
\definecolor{ao(english)}{rgb}{0.0, 0.5, 0.0}

\begin{document}
\pagestyle{headings}
\mainmatter
\def\ECCVSubNumber{5091}  

\title{Unsupervised Lifelong Person Re-identification via Contrastive Rehearsal} 

\titlerunning{Unsupervised Lifelong Person ReID via Contrastive Rehearsal}
%
\author{Hao Chen\inst{1,2} \and
Benoit Lagadec\inst{2} \and
Francois Bremond\inst{1}}
\authorrunning{H. Chen et al.}
%
\institute{Inria, Université Côte d'Azur, 06902 Valbonne, France\\
\email{\{hao.chen, francois.bremond\}@inria.fr} \and
European Systems Integration, 06110 Le Cannet, France\\
\email{benoit.lagadec@esifrance.net}}
\maketitle

\begin{abstract}
Existing unsupervised person re-identification (ReID) methods focus on adapting a model trained on a source domain to a fixed target domain. However, an adapted ReID model usually only works well on a certain target domain, but can hardly memorize the source domain knowledge and generalize to upcoming unseen data. In this paper, we propose unsupervised lifelong person ReID, which focuses on continuously conducting unsupervised domain adaptation on new domains without forgetting the knowledge learnt from old domains. To tackle unsupervised lifelong ReID, we conduct a contrastive rehearsal on a small number of stored old samples while sequentially adapting to new domains. We further set an image-to-image similarity constraint between old and new models to regularize the model updates in a way that suits old knowledge. We sequentially train our model on several large-scale datasets in an unsupervised manner and test it on all seen domains as well as several unseen domains to validate the generalizability of our method. Our proposed unsupervised lifelong method achieves strong generalizability, which significantly outperforms previous lifelong methods on both seen and unseen domains. Code will be made available at \url{https://github.com/chenhao2345/UCR}.

\end{abstract}

\section{Introduction}
\label{sec:intro}

Person re-identification (ReID) targets at matching a person of interest across non-overlapping cameras. Although significant improvement has been witnessed in both supervised~\cite{zheng2019joint,He_2021_ICCV} and unsupervised~\cite{ge2020self,Chen_2021_ICE} person ReID, traditional methods only consider the performance of a single fixed target domain. In the single target domain scenario, people usually assume that all training data is available before training and deploying a ReID model. However, a real-world video monitoring system can record new data every day and from new locations, when new cameras are added into an existing system. How to adapt a model to new data without  catastrophic forgetting on old knowledge has become a key point for training a generalizable and robust ReID model.


Towards a generalizable ReID model, \textit{lifelong person ReID}~\cite{pu_cvpr2021,Wu2021GeneralisingWF} has been recently proposed to incrementally accumulate domain knowledge from several seen datasets. Lifelong person ReID is related to incremental (or continuous) learning~\cite{Li2018LearningWF}, which aims at incrementally adding new classes or new domain knowledge into an existing model. As training and test sets in person ReID have non-overlapping identities, lifelong person ReID is defined as a domain-incremental learning task. A lifelong trained model has proven to be effective on every seen domain, as well as on unseen domains. However, previous lifelong person ReID relies on supervised cross-domain fine-tuning. When new data is recorded every day, people have to annotate new data manually before deployment, which is cumbersome and time-consuming. Replacing supervised cross-domain fine-tuning with unsupervised domain adaptation can maximally enhance the flexibility of a lifelong person ReID algorithm in real-world deployments.

\begin{figure}[t]
\centering
   \includegraphics[width=0.5\linewidth]{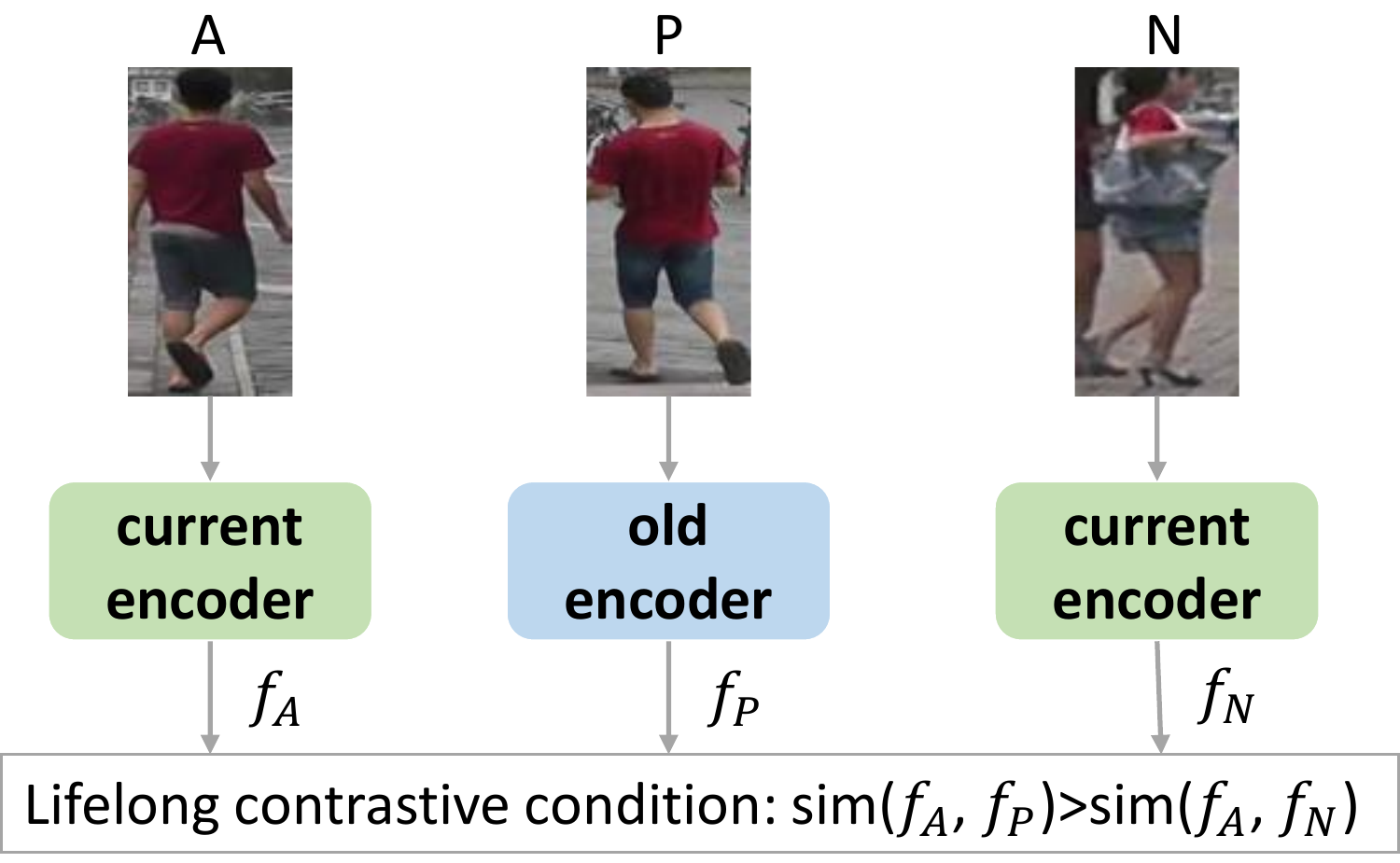}
   \caption{Unsupervised lifelong contrastive condition: the representation of an anchor (A) should always be more similar to a pseudo positive (P) than a pseudo negative (N), even though $f_A$ and $f_N$ are encoded with same current domain knowledge while $f_P$ is encoded with old domain knowledge.}
\label{fig:figure1}
\end{figure}

In this paper, we propose a new \textit{unsupervised lifelong person ReID} task to simultaneously explore 1) the possibility of training a generalizable model on incrementally added domains without human supervision and 2) the possibility of mitigating catastrophic forgetting problem neglected in traditional unsupervised person ReID. To train unsupervised lifelong person ReID, we have to consider learning unsupervised new domain representations and maintaining old domain knowledge simultaneously. In such context, we incorporate pseudo label based contrastive learning and rehearsal-based incremental learning into an \textit{unsupervised contrastive rehearsal} (UCR) method, which tackles the forgetting problem during the unsupervised representation learning. 

In our proposed UCR, a small number of old domain samples and their corresponding cluster prototypes are stored in long-term memory buffers. While adapting a model to a new domain, rehearsing stored old domain samples helps to prevent forgetting old knowledge. Given a frozen old domain model and a current domain model, we set a lifelong contrastive condition in Fig.~\ref{fig:figure1}: an old sample should always be closer to its pseudo positives than any pseudo negatives regardless of domain changes. Based on this condition, we try to retrieve positive pairs across different domain knowledge, which effectively mitigates forgetting in an unsupervised manner. Moreover, given a batch of old samples, the image-to-image similarity calculated by the old domain model and the new domain model should be consistent. We thus regularize the image-to-image representation relationship between old and new domain models, so that the new domain model can be updated in a way that suits old knowledge.

To summarize, our contributions are: 1) We propose a challenging but practical unsupervised lifelong person ReID task, which targets at incrementally learning a generalizable ReID model without human supervision. 2) We propose a contrastive rehearsal method and a representation relationship constraint to mitigate the forgetting problem in unsupervised lifelong person ReID. 3) Extensive experiments on both seen datasets and unseen datasets validate the effectiveness of our proposed method in unsupervised lifelong person ReID.

\section{Related Work}
\paragraph{Person ReID.} Depending on the number of training/test domains and availability of human annotation, recent person ReID research is conducted under different settings. As the most studied setting, supervised person ReID~\cite{chen2019abd,zheng2019joint,Luo_2019_CVPR_Workshops,He_2021_ICCV} has shown impressive performance on large-scale datasets thanks to deep learning methods and human annotation. However, as a fine-grained retrieval task, a ReID model trained on one domain is hard to generalize to other domains. Unsupervised domain adaptation~\cite{ge2020mutual,ge2020self,Dai_2021_ICCV} and fully unsupervised ReID~\cite{Wang2021camawareproxies,Chen_2021_ICE} are proposed to adjust a ReID model to a target domain with unlabeled target domain images. On the other hand, domain generalization ReID~\cite{Song2019GeneralizablePR,Jin_2020_CVPR,dai2021generalizable} is proposed to jointly train multiple labeled domains, in order to learn a generalizable model that can still work on unseen domains. But in most real-world cases, it is hard to prepare all training data in advance. Instead, new domain data can be recorded when time and season change or a new camera is installed. Supervised lifelong person ReID~\cite{pu_cvpr2021,Wu2021GeneralisingWF} is thus proposed to learn incrementally added new domains. However, continuously annotating new domains can be a cumbersome task for ReID system administrators. In this paper, we introduce unsupervised lifelong person ReID to maximally improve the flexibility of lifelong person ReID in the real-world deployments. We propose a contrastive rehearsal method to mitigate the catastrophic forgetting during the sequential unsupervised domain adaptation. Our proposed method is mainly related to contrastive learning and lifelong learning.

\paragraph{Contrastive learning.} The main idea of contrastive learning is to maximize the representation similarity between a positive pair composed of differently augmented views of a same image, so that a model can understand the augmented variance is noise. While attracting a positive pair, some contrastive methods also consider other images as negatives and push away negatives stored in a memory bank~\cite{Wu2018UnsupervisedFL,He_2020_CVPR} or in a large mini-batch~\cite{chen2020simple}. 
Contrastive methods show great performance in unsupervised representation learning, which makes it a main approach in unsupervised person ReID. Based on clustering generated pseudo labels, state-of-the-art unsupervised person ReID methods build positive pairs with cluster centroids~\cite{ge2020self}, camera-aware centroids~\cite{Wang2021camawareproxies}, mini-batch hardest positives~\cite{Chen_2021_ICE} and generated positive views~\cite{Chen_2021_CVPR}. However, all these methods are designed for single-domain unsupervised ReID, in which only representations from a single domain are contrasted. Differently, we propose to mitigate the catastrophic forgetting by contrasting representations across current and old domain knowledge.

\paragraph{Lifelong learning.} Lifelong (also called incremental or continuous) learning aims at learning new classes or new domains without forgetting old knowledge. Several approaches have been proposed to address the forgetting problem in lifelong learning. One of the most intuitive approaches is rehearsal (also called recall or replay) learning , in which a small number of old data, such as raw images~\cite{Rebuffi2017iCaRLIC,Castro2018End,cha2021co2l} or feature vectors~\cite{iscen2020memory}, are stored to remind the new model of old knowledge. Another approach~\cite{yoon2018lifelong,chaudhry2018riemannian} consists in regularizing model updates on new data in a way that does not contradict the old knowledge. The third approach is mainly based on knowledge distillation~\cite{Li2018LearningWF,aljundi2018memory,douillard2021plop}, which considers the old model as a teacher for the new model. Two supervised lifelong person ReID methods have been recently proposed. In AKA~\cite{pu_cvpr2021}, authors distill old knowledge via a learnable knowledge graph to the new domain model. GwFReID~\cite{Wu2021GeneralisingWF} stores 2 images per old domain identity for rehearsal and regularizes coherence between old domain and new domain models in both representation and classifier prediction levels. On the other hand, several attempts have been made in unsupervised lifelong adaptation, such as setting gradient regularization~\cite{tang2021gradient} in contrastive learning and consolidating the internal distribution~\cite{rostami2021lifelong}. However, general lifelong adaptation~\cite{tang2021gradient,rostami2021lifelong} has fixed classes across different domains, which are not suitable for lifelong ReID that has to learn fine-grained identity representations from totally different classes across domains.

\section{Methodology}
\begin{figure*}
\centering
   \includegraphics[width=1\linewidth]{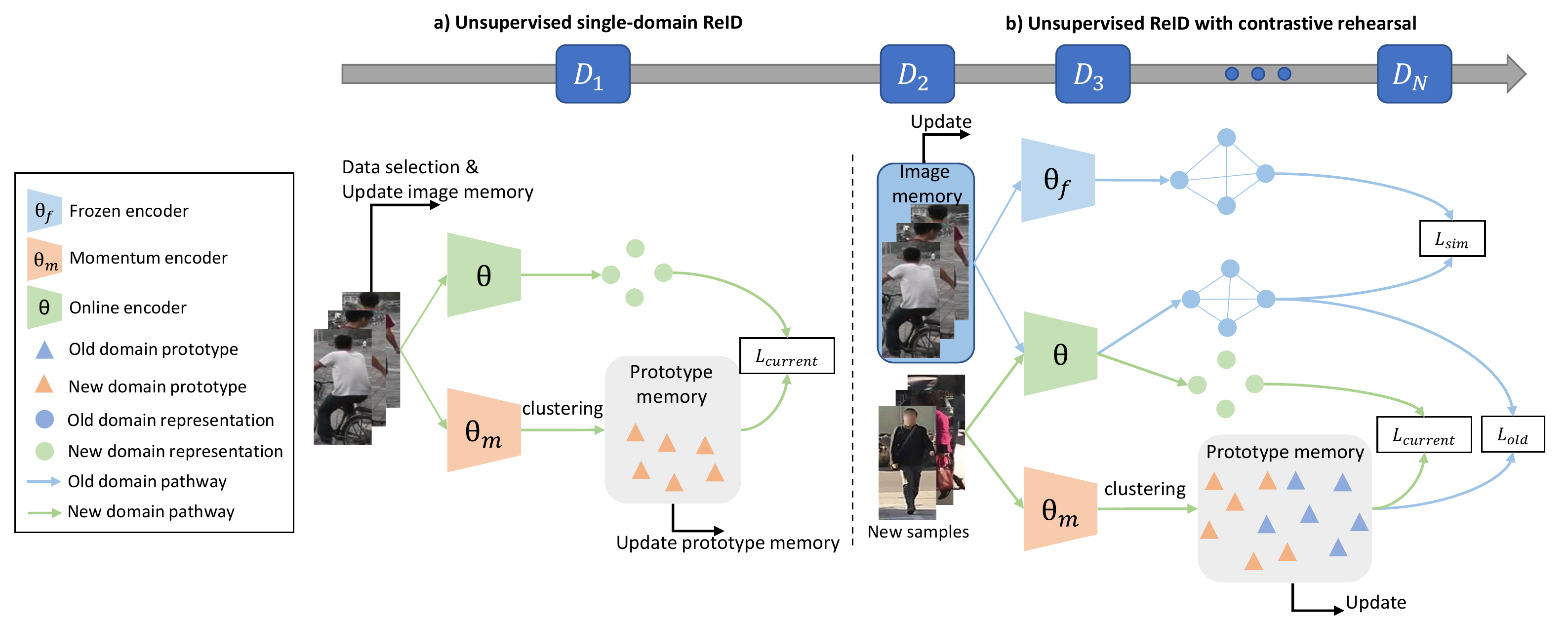}
   \caption{General architecture of our proposed method. On the first domain $D_{1}$, we only follow the new domain pathway (\textcolor{green}{$\rightarrow$}) to conduct current domain contrastive learning $\mathcal{L}_{current}$. On the following domains $D_{2}\to ... \to D_{N}$, we follow both new domain pathway (\textcolor{green}{$\rightarrow$}) and old domain rehearsal pathway(\textcolor{blue}{$\rightarrow$}) to conduct image-to-prototype contrastive learning $\mathcal{L}_{current}$ and $\mathcal{L}_{old}$ , as well as similarity constraint $\mathcal{L}_{sim}$. }
\label{fig:figure2}
\end{figure*}
\subsection{Overview}
Given a stream of $N$ person ReID datasets, unsupervised lifelong person ReID aims at learning a generalizable model via sequential unsupervised learning on the training set of each domain $D_{1}\to D_{2}\to ... \to D_{N}$. An unsupervised lifelong person ReID pipeline can be defined as one fully unsupervised ReID step on $D_{1}$ followed by several unsupervised domain adaptation steps on $D_{2}, ..., D_{N}$. After the whole pipeline, the adapted model is tested respectively on the test set of each seen domain, as well as on multiple unseen domains.

We present a new UCR method for lifelong person ReID. The general architecture of UCR is illustrated in Fig.~\ref{fig:figure2}. In the first step, we train a model on unlabeled images in the domain $D_{1}$ through a fully unsupervised ReID method. We follow state-of-the-art fully unsupervised ReID methods to use pseudo label based contrastive learning as a baseline, which considers both an online encoder and a momentum encoder, where the momentum encoder serves as a knowledge collector that gradually accumulates knowledge of each seen domain. To stabilize pseudo labels with momentum representations, the momentum encoder (weights noted as $\theta_m$) is updated by exponential moving averaged weights of the online encoder (weights noted as $\theta$):
\begin{equation}
\theta_m^t =\alpha\theta_m^{t-1}+(1-\alpha)\theta^t
\label{equ:ema}
\end{equation}
where the hyper-parameter $\alpha$ controls the update speed of the momentum encoder. $t$ and $t-1$ refer respectively to the current and last iteration. We extract all image representations with the stable momentum encoder and generate corresponding pseudo labels with a density-based clustering algorithm DBSCAN~\cite{Ester1996ADA}. Based on the clustered pseudo labels, we build cluster prototypes for a current domain image-to-prototype contrastive loss $\mathcal{L}_{current}$ (described in Section \ref{sec:Centroid-based Contrastive Baseline}) on the current domain $D_{1}$. To mitigate the forgetting, we build an image memory and a prototype memory to store old domain images and cluster prototypes for rehearsal. The memory buffers are updated after training each domain.

In the following adaptation step on domain $D_i$, we continue using the prototype contrastive loss $\mathcal{L}_{current}$ on new domain images to learn new knowledge, wherein pseudo labels are generated on the current domain images in the same way as the first step. On the other hand, we freeze the momentum encoder from the last step ($\theta_m^{D_{i-1}} \to\theta_f^{D_i}$) as an old knowledge expert model. Based on our lifelong contrastive condition that an old sample encoded by the new model $\theta$ should be close to its prototype, we formulate an old domain rehearsal loss $\mathcal{L}_{old}$ in Section~\ref{sec:Contrastive Rehearsal}. We further set an image-to-image similarity constraint (Section~\ref{sec:Image-to-image Similarity Distillation}) to regularize the model updates during the continuous adaptation.

The overall unsupervised lifelong loss is defined as:
\begin{equation}
\mathcal{L}_{overall} = \mathcal{L}_{current}+\mathcal{L}_{old} +\lambda_{sim} \mathcal{L}_{sim}
\label{equ:overall}
\end{equation}

\subsection{Current domain contrastive baseline}
\label{sec:Centroid-based Contrastive Baseline}
Inside an unsupervised lifelong ReID pipeline, our model incrementally learns new knowledge on a current domain $D^{c}=\{(x_{1}^{c}, y_{1}^{c}), ..., (x_{N_{D_{i}}}^{c}, y_{N_{D_{i}}}^{c})\}$ where $N_{D_{i}}$ is the number of images and $y$ is the clustered pseudo label of $x$. For a current domain image $x_{i}^{c}$, $f(x_{i}^{c}|\theta)$ and $f(x_{i}^{c}|\theta_m)$ denote respectively the online and the momentum representations. The prototype of a cluster $a$ is defined as the averaged momentum representations of all the samples with a same pseudo label $y_{a}$:
\begin{equation}
   p^{c}_{a} = \frac{1}{N_{a}}\sum_{x_{i}^{c} \in y_{a}} f(x_{i}^{c}|\theta_m)
\label{equa:cluster centroid}
\end{equation}
When $x_{i}^{c}$ belongs to the cluster $a$, a cluster prototype contrastive loss~\cite{ge2020self} can be defined as:
\begin{equation}
   \mathcal{L}_{cluster} = \mathop{\mathbb{E}}[-\log{\frac{\exp{(f(x_{i}^{c}|\theta) \cdot p_{a}^c/\tau_p)}}{\sum\nolimits_{j=1}^{|P_{}^{c}|}\exp{(f(x_{i}^{c}|\theta) \cdot p_{j}/\tau_p)}}}]
\label{equa:cluster}
\end{equation}
where $|P^{c}|$ is the total number of clusters in the current domain and $\tau_p$ is a temperature hyper-parameter.

$\mathcal{L}_{cluster}$ makes samples in a cluster converge to a common prototype and get far away from other clusters. As a ReID dataset is usually recorded across different cameras, minimizing the intra-cluster variance from different camera style has proven to be effective in person ReID~\cite{Wang2021camawareproxies}. Supposing the current domain is recorded by $N_C$ cameras $\mathcal{C}=\{c_1, ..., c_{N_C}\}$, an intra-cluster camera prototype is defined as the averaged momentum representation of all the samples with a same pseudo label $y_{a}$ that are recorded from a same camera $c_{b}$:
\begin{equation}
p^{c}_{ab} = \frac{1}{N_{ab}}\sum_{x^{c}_{i} \in y_{a} \cap x^{c}_{i} \in c_{b}}^{}  f(x_{i}^{c}|\theta_m)
\label{equa:camera centroid}
\end{equation}
When $x_{i}^{c}$ has a pseudo label $y_a$ and is recorded from $c_{b}$, a camera prototype contrastive loss can be defined as:
\begin{equation} \small
   \mathcal{L}_{cam} = \mathop{\mathbb{E}}[- \frac{1}{N_C} \sum_{j \in \mathcal{C}} \log{\frac{\exp{(f(x_{i}^{c}|\theta) \cdot p_{aj}^{c}/\tau_{c})}}{\sum_{k=1}^{N_{neg}+1}\exp{(f(x_{i}^{c}|\theta) \cdot p_{k}^c/\tau_{c})}}}]
\label{equa:camera}
\end{equation}
where $\tau_{c}$ is a camera contrastive temperature hyper-parameter. $N_{neg}$ hardest negative camera prototypes from the current domain are selected to enhance the model discriminability. $\mathcal{L}_{cam}$ maximizes the similarity between a representation and all the camera prototypes within a same cluster to reduce intra-cluster variance. 

An overall loss on the current domain combines Eq. (\ref{equa:cluster}) and (\ref{equa:camera}) with a balancing hyper-parameter $\lambda_{cam}$:
\begin{equation}
\mathcal{L}_{current} = \mathcal{L}_{cluster}+\lambda_{cam} \mathcal{L}_{cam}
\label{equ:current}
\end{equation}
\textbf{Remark.} By filtering out intra-cluster variance from different camera styles, $\mathcal{L}_{cam}$ purifies the current domain knowledge before being accumulated into the model. However, $\mathcal{L}_{cam}$ relies on camera labels, which make our method more ReID-specific. In fact, $\mathcal{L}_{cam}$ could be replaced with other techniques that do not require camera labels, such as contrasting mini-batch hardest positives~\cite{Chen_2021_ICE} (see Supplementary Materials). 

\subsection{Old domain contrastive rehearsal}
\label{sec:Contrastive Rehearsal}
At the end of each step, we store all the cluster prototypes into a prototype memory and $K_{mem}$ images per cluster into an image memory. To reduce pseudo label noise for contrastive rehearsal, for each cluster, we select $K_{mem}$ images that have highest cosine similarity with the cluster prototype as reliable images to be stored. For the prototype memory, only general cluster prototypes but not camera-aware prototypes are stored to keep the memory buffer in a reasonable size.

At the beginning of adaptation on the domain $D_{i}$, the stored old domain cluster prototypes $P^o=\{P^{D_1}, ..., P^{D_{i-1}}\}$ are concatenated with current cluster prototypes $P^c=\{P^{D_i}\}$ (encoded by the current momentum encoder $\theta_m$ at the beginning of each epoch) to update the prototype memory $P=P^{o}\cup P^{c}$. Given an old sample $x_{i}^{o}$ of identity $y_a$, if the old knowledge is well maintained in the current model, the online representation $f(x_{i}^o|\theta_{})$ encoded by the current domain online encoder $\theta$ should have the highest similarity score with the stored $p_{a}^o$ encoded by the old domain encoder from the prototype memory. Thus, we construct an old domain contrastive rehearsal loss on stored old samples to remind the current model of old domain knowledge by maximizing the similarity between the old domain image representation $f(x_{i}^o|\theta_{})$ and the stored corresponding prototype $p_{a}^o$, while minimizing the similarity between $f(x_{i}^o|\theta_{})$ and other prototypes in the prototype memory:
\begin{equation}
   \mathcal{L}_{old} = \mathop{\mathbb{E}}[-\log{\frac{\exp{(f(x_{i}^o|\theta_{}) \cdot p_{a}^o/\tau_p)}}{\sum\nolimits_{j=1}^{|P|}\exp{(f(x_{i}^o|\theta_{}) \cdot p_{j}/\tau_p)}}}]
\label{equa:old}
\end{equation}
where $|P|$ is the number of cluster prototypes in the prototype memory and $\tau_p$ is the prototype temperature hyper-parameter same as Eq. (\ref{equa:cluster}). As a cluster prototype (the averaged representation of all cluster samples) contains generic information of a cluster, the prototype memory enables the current model to have access to generic old domain cluster information without storing all the images.

\subsection{Image-to-image Similarity Constraint}
\label{sec:Image-to-image Similarity Distillation}
Technically, person ReID is a representation similarity ranking problem, in which the objective is to have high similarity scores between positive pairs and low similarity scores between negative pairs. However, when a model is adapted into a new domain, the similarity relationship between old domain samples could be affected by the new domain knowledge. As the similarity relationship between same images should be consistent before and after a domain adaptation step, we propose an image-to-image similarity constraint loss that regularizes the similarity relationship updates in a way that does not contradict the old knowledge.
As the frozen old model $\theta_{f}$ from the last domain can be regarded as an expert on the old domain, the similarity relationship calculated by the frozen model can be regarded as a reference for regularizing the current model $\theta_{}$ updates.

Given a mini-batch of old images $\{x_{1}^{o}, ..., x_{N_{bs}}^{o}\}$ where $N_{bs}$ is the batch size, the image-to-image similarity distribution can be calculated with a softmax function on the cosine similarity between each image pair in the mini-batch. The image-to-image similarity between two old images $x_{i}^{o}$ and $x_{j}^{o}$ is calculated with both the online encoder $\theta_{}$ and the momentum encoder $\theta_{m}$:
\begin{equation}
   P_{i,j} = \frac{\exp{(<f(x_{i}^{o}|\theta_{}) \cdot f(x^{o}_{j}|\theta_{m})>/\tau_{s})}}{\sum\nolimits_{k=1}^{N_{bs}}\exp{(<f(x_{i}^{o}|\theta_{}) \cdot f(x^{o}_{k}|\theta_{m})>/\tau_{s})}}
\label{equa:P}
\end{equation}
where $< \cdot >$ denotes the normalized cosine similarity and $\tau_{s}$ is a similarity temperature hyper-parameter. 

For the same mini-batch, we calculate the image-to-image similarity distribution with the frozen old model $\theta_{f}$ as a reference for the constraint. The reference similarity between two old domain images $x_{i}^{o}$ and $x_{j}^{o}$ is:
\begin{equation}
Q_{i,j} = \frac{\exp{(<f(x_{i}^{o}|\theta_{f}) \cdot f(x^{o}_{j}|\theta_{f})>/\tau_{s})}}{\sum\nolimits_{k=1}^{N_{bs}}\exp{(<f(x_{i}^{o}|\theta_{f}) \cdot f(x^{o}_{k}|\theta_{f})>/\tau_{s})}}
\label{equa:Q}
\end{equation}

We formulate an image-to-image similarity constraint loss with a Kullback-Leibler (KL) Divergence between the two distributions:
\begin{equation}
   \mathcal{L}_{sim} = \mathcal{D}_{KL}(P||Q)
\label{equa:consistency}
\end{equation}
By minimizing $\mathcal{L}_{sim}$, we encourage the similarity relationship $P$ calculated with current domain knowledge to be consistent with that calculated with old domain knowledge $Q$.

\textbf{Remark.} Here, the similarity $P_{i,j}$ is calculated in online/momentum ($\theta/\theta_{m}$) format, which is the similarity between current online representations and accumulated momentum representations. Such online/momentum similarity encourages the online encoder $\theta$ updates in a way that is consistent with the accumulated momentum encoder $\theta_m$, which is better than only consider online/online ($\theta/\theta$) similarity.

\section{Experiment}

\subsection{Datasets and Evaluation Protocols}
As DukeMTMC-reID dataset~\cite{ristani2016MTMC} has been taken down from the website, we do not follow previous lifelong ReID benchmarks~\cite{Wu2021GeneralisingWF,pu_cvpr2021}. Instead, we set up a new lifelong person ReID benchmark, which contains 3 seen datasets for domain-incremental training and 9 unseen datasets for generalizability evaluation, as shown in Table~\ref{tab:datasets}. Compared with previous supervised lifelong ReID benchmarks~\cite{Wu2021GeneralisingWF,pu_cvpr2021} with DukeMTMC-reID, our benchmark contains less seen domains but more unseen domains, which can better evaluate the model generalizability.  
\begin{table}[t]
\centering
\scalebox{0.8}{
\begin{tabular}
{c|ccccc}
\hline
Type& Dataset  & \#img & \#train id & \#train img & \#test id \\
\hline
\multirow{3}{*}{Seen} & Market~\cite{Zheng2015ScalablePR} & 36036 & 751& 12936 & 750\\
 & Cuhk-Sysu~\cite{Xiao2017JointDA} & 23435 & 5532& 15088 & 2900\\
 & MSMT17~\cite{wei2018person} & 124068 & 1041 & 32621 & 3060\\
\hline
\multirow{9}{*}{Unseen} & VIPeR~\cite{Gray2008ViewpointIP}&1264&-&-&316\\
 & PRID~\cite{hirzer11}&1134&-&-&649\\
 & GRID~\cite{Loy2009MulticameraAC}&1275&-&-&125\\
 & iLIDS~\cite{Zheng2009AssociatingGO}&476&-&-&60\\
 & CUHK01~\cite{Li2012HumanRW}&1942&-&-&486\\
 & CUHK02~\cite{Li2013LocallyAF}&7264&-&-&239\\
 & SenseReID~\cite{Zhao2017SpindleNP}&4428&-&-&1718\\ 
 & CUHK03~\cite{Li2014DeepReIDDF}&14097&-&-&100\\
 & 3DPeS~\cite{3dpes}&1012&-&-&96\\
\hline
\end{tabular}}
\caption{Dataset statistics. Unseen domains are only used for testing.}
\label{tab:datasets}
\end{table}
\paragraph{Seen datasets.} We use three large-scale datasets as seen domains, including Market, Cuhk-Sysu and MSMT17. We use the ReID version of Cuhk-Sysu dataset, in which each person bounding box is cropped from raw images using the ground-truth annotation. We formulate a 3-step domain-incremental training in the order Market$\to$Cuhk-Sysu$\to$MSMT17. Our model is trained sequentially on the training set of each seen dataset and is tested on the test set of each dataset after the final step. Cumulative Matching Characteristics (CMC) at Rank1 accuracy and mean Average Precision (mAP) are used in our experiments. 

\paragraph{Unseen dataset.} We use 9 person ReID datasets to maximally evaluate the model generalizability on different unseen domains, including VIPeR, PRID, GRID, iLIDS, CUHK01, CUHK02, SenseReID, CUHK03 and 3DPeS. These 9 datasets cover all the unseen domains that are considered in previous supervised lifelong ReID methods~\cite{Wu2021GeneralisingWF,pu_cvpr2021} and domain generalizable ReID methods~\cite{Song2019GeneralizablePR}. We use the traditional training/test split on CUHK03 dataset. Rank1 accuracy and mAP results are respectively reported on the test set of each unseen domain after the final step. 

\subsection{Implementation details}
\paragraph{Training.} Our method is implemented under Pytorch~\cite{PyTorch_NEURIPS2019} framework. The total training time with 4 Nvidia 1080Ti GPUs is around 6 hours. We use an ImageNet \cite{Russakovsky2015ImageNetLS} pre-trained ResNet50 \cite{he2016deep} as our backbone network. We resize all images to 256$\times$128 and augment images with random horizontal flipping, cropping, Gaussian blurring and erasing \cite{Zhong2020RandomED}. At each step (domain), we train our framework 30 epochs with 400 iterations per epoch using a Adam~\cite{Adam_optimization} optimizer with a weight decay rate of 0.0005. The learning rate is set to 0.00035 with a warm-up scheme in the first 10 epochs. No learning rate decay is used in the training. Pseudo labels on the current domain are updated on re-ranked Jaccard distance~\cite{zhong2017re} at the beginning of each epoch with a DBSCAN~\cite{Ester1996ADA}, in which the minimum cluster sample number is set to 4 and the distance threshold is set to 0.55. The momentum encoder is updated with a momentum hyper-parameter $\alpha=0.999$. Following~\cite{Chen_2021_ICE}, we set $\tau_p=0.5$, $\tau_c=0.07$, $N_{neg}=50$ and $\lambda_{cam}=0.5$ in Eq. (\ref{equa:camera}) in the baseline. We use a grid search on $\tau_s$ and $\lambda_{sim}$, which are presented in Section \ref{sec:Parameter analysis}. After the whole training, only the momentum encoder is saved for inference. We provide more details of our algorithm in Supplementary Materials.

\paragraph{Mini-batch composition.} To balance the model ability on old domains and a current domain, we separately take a mini-batch of current domain images and a mini-batch of old domain images of a same batch size, which is set to 32 in our experiments. Furthermore, we use a random identity sampler to construct mini-batches to handle the imbalanced images of different identities. Following the clustering setting on the current domain (each cluster has at least 4 neighbors), the 32 current domain images are composed of 8 identities and 4 images per identity. Following the supervised lifelong ReID method~\cite{Wu2021GeneralisingWF}, we set $K_{mem}=2$ to store 2 images per cluster. The 32 old domain images are thus composed of 16 identities and 2 images per identity.

\begin{table}[t]
\scalebox{0.8}{
\begin{tabular}{l|c|c|cc|cc|cc|cc}
\hline
\multicolumn{11}{c}{Training order: Market$\to$Cuhk-Sysu$\to$MSMT17}\\\hline
\multirow{2}{*}{Method} & Memory & \multirow{2}{*}{Type} & \multicolumn{2}{c}{Market} & \multicolumn{2}{|c}{Cuhk-Sysu} & \multicolumn{2}{|c}{MSMT17} & \multicolumn{2}{|c}{Average} \\ \cline{4-11}
\multicolumn{1}{c}{} &\multicolumn{1}{|c|}{(image per id)}&\multicolumn{1}{c|}{}& \multicolumn{1}{c}{mAP} & \multicolumn{1}{c|}{Rank1} &\multicolumn{1}{c}{mAP} & \multicolumn{1}{c|}{Rank1} &\multicolumn{1}{c}{mAP} & \multicolumn{1}{c|}{Rank1} &\multicolumn{1}{c}{mAP} & \multicolumn{1}{c}{Rank1}  \\ 
\hline
SpCL~\cite{ge2020self}&0&U&18.8&39.2&70.9&74.4&10.5&24.8&33.4&46.1\\
ICE~\cite{Chen_2021_ICE}&0&U&29.0&60.4&72.5&76.3&21.8&49.0&41.1&61.9\\\hline
BL+LwF~\cite{Li2018LearningWF}&0&UL&34.3&67.7&73.3&76.8&23.5&52.5&43.7&65.6\\
BL+SPD~\cite{Tung2019SimilarityPreservingKD}&0&UL&33.4&65.3&74.9&78.1&27.5&58.9&45.3&67.4\\
BL+iCaRL~\cite{Rebuffi2017iCaRLIC}&2&UL&38.6&67.7&80.6&83.0&26.3&56.3&48.5&69.0\\
BL+C$o^2$L~\cite{cha2021co2l}&2&UL&43.5&72.7&78.5&81.0&30.4&61.5&50.8&71.7\\
BL+\textbf{UCR}&2&UL&57.6&83.0&83.2&85.6&25.4&54.1&\textbf{55.4}&\textbf{74.3}\\
\hline
AKA~\cite{pu_cvpr2021}&0&SL&57.7&78.6&77.0&80.0&9.2&21.5&48.0&60.0\\
BL(GT)+LwF~\cite{Li2018LearningWF}&0&SL&39.6&68.9&80.7&83.7&44.1&72.0&54.8&74.9\\
BL(GT)+SPD~\cite{Tung2019SimilarityPreservingKD}&0&SL&37.7&66.5&80.8&83.0&41.5&69.7&53.3&73.1\\
BL(GT)+iCaRL~\cite{Rebuffi2017iCaRLIC}&2&SL&32.8&60.6&85.4&87.7&43.5&70.8&53.9&73.1\\
BL(GT)+C$o^2$L~\cite{cha2021co2l}&2&SL&48.6&74.5&84.1&86.3&45.2&72.3&59.3&77.7\\
BL(GT)+\textbf{UCR}&2&SL&59.3&82.7&88.3&90.0&40.8&67.5&\textbf{62.8}&\textbf{80.1}\\
\hline
\end{tabular}}
\centering
\caption{Seen-domain results (\%) of unsupervised single-domain (U), supervised lifelong (SL) and unsupervised lifelong (UL) methods. `BL' denotes our current domain baseline. `BL(GT)' refers to replacing pseudo labels with ground truth labels.}
\label{table:seen domains results}
\end{table}

\begin{table}[t]
\scalebox{0.58}{
\begin{tabular}{l|c|c|cc|cc|cc|cc|cc|cc|cc|cc|cc|cc}
\hline
\multicolumn{23}{c}{Training order: Market$\to$Cuhk-Sysu$\to$MSMT17}\\\hline
\multirow{2}{*}{Method} & Memory&\multirow{2}{*}{Type}& \multicolumn{2}{|c}{VIPeR} & \multicolumn{2}{|c}{PRID} & \multicolumn{2}{|c}{GRID} & \multicolumn{2}{|c}{iLIDS} & \multicolumn{2}{|c}{CUHK01}& \multicolumn{2}{|c}{CUHK02}& \multicolumn{2}{|c}{SenseReID}& \multicolumn{2}{|c}{CUHK03}& \multicolumn{2}{|c}{3DPeS}& \multicolumn{2}{|c}{Average} \\ \cline{4-23}
\multicolumn{1}{c}{}&\multicolumn{1}{|c|}{(per id)}&\multicolumn{1}{|c}{}& \multicolumn{1}{|c}{mAP} & \multicolumn{1}{c|}{R1} &\multicolumn{1}{c}{mAP} & \multicolumn{1}{c|}{R1} &\multicolumn{1}{c}{mAP} & \multicolumn{1}{c|}{R1} &\multicolumn{1}{c}{mAP} & \multicolumn{1}{c|}{R1} &\multicolumn{1}{c}{mAP} & \multicolumn{1}{c|}{R1} &\multicolumn{1}{c}{mAP} & \multicolumn{1}{c|}{R1} &\multicolumn{1}{c}{mAP} & \multicolumn{1}{c|}{R1} &\multicolumn{1}{c}{mAP} & \multicolumn{1}{c|}{R1} &\multicolumn{1}{c}{mAP} & \multicolumn{1}{c|}{R1} &\multicolumn{1}{c}{mAP} & \multicolumn{1}{c}{R1}  \\ 
\hline
SpCL~\cite{ge2020self}&0&U&31.7&22.8&9.2&4.0&10.6&5.6&58.7&48.3&45.2&44.9&40.2&37.9&28.1&22.1&8.1&22.2&35.5&43.1&29.7&27.9\\
ICE~\cite{Chen_2021_ICE}&0&U&35.7&25.9&39.0&29.0&20.6&14.4&71.4&61.7&60.6&60.0&48.2&45.8&33.6&27.9&17.3&29.7&48.5&55.4&41.7&38.9\\\hline
BL+LwF~\cite{Li2018LearningWF}&0&UL&41.2&32.3&49.0&37.0&31.9&23.2&76.8&66.7&63.4&62.0&54.1&52.3&37.6&31.2&21.7&35.0&53.1&59.4&47.7&44.3\\
BL+SPD~\cite{Tung2019SimilarityPreservingKD}&0&UL&41.3&30.1&29.7&22.0&27.4&19.2&79.2&71.7&67.5&67.1&55.4&55.9&39.7&33.1&21.0&37.9&50.8&57.9&45.8&43.9\\
BL+iCaRL~\cite{Rebuffi2017iCaRLIC}&2&UL&45.9&35.1&48.6&39.0&32.5&22.4&78.9&71.7&66.2&66.9&57.5&55.6&43.8&36.4&24.1&41.6&56.7&64.4&50.5&48.1\\
BL+C$o^2$L~\cite{cha2021co2l}&2&UL&47.7&37.0&51.1&40.0&28.4&20.0&80.6&73.3&70.5&71.0&62.3&60.5&44.0&36.5&29.8&43.9&60.0&66.8&52.7&49.9\\
BL+\textbf{UCR}&2&UL&47.7&37.0&55.5&47.0&40.6&31.2&85.3&81.7&69.8&68.8&68.0&65.3&47.0&39.5&33.0&48.0&64.9&71.3&\textbf{56.8}&\textbf{54.4}\\
\hline

AKA~\cite{pu_cvpr2021}&0&SL&37.9&28.8&31.0&21.0&24.0&15.2&70.6&60.0&54.1&53.3&47.2&43.9&34.8&28.1&19.5&19.3&43.9&56.3&40.3&36.2\\
GwFReID*~\cite{Wu2021GeneralisingWF}&2&SL&-&43.2&-&-&-&-&-&69.5&-&-&-&-&-&-&-&40.2&-&64.9&-&-\\
BL(GT)+LwF~\cite{Li2018LearningWF}&0&SL&51.5&40.5&41.8&33.0&26.8&20.8&79.1&71.7&74.4&75.8&62.2&61.7&44.0&37.4&29.8&47.4&55.3&60.9&51.7&49.9\\
BL(GT)+SPD~\cite{Tung2019SimilarityPreservingKD}&0&SL&48.7&37.7&25.5&15.0&23.6&16.8&81.9&75.0&70.9&71.8&60.9&60.5&44.2&36.8&26.9&45.7&54.6&65.8&48.6&47.2\\
BL(GT)+iCaRL~\cite{Rebuffi2017iCaRLIC}&2&SL&52.4&41.5&45.9&37.0&32.4&22.4&82.0&75.0&69.1&70.3&62.5&61.1&47.0&39.7&33.3&51.0&57.2&63.4&53.5&51.3\\
BL(GT)+C$o^2$L~\cite{cha2021co2l}&2&SL&56.3&46.2&52.3&41.0&28.4&20.8&83.5&76.7&77.5&78.3&67.0&65.1&48.9&41.2&37.1&54.5&60.5&67.8&56.8&54.6\\
BL(GT)+\textbf{UCR}&2&SL&57.7&47.5&56.0&44.0&40.6&31.2&87.9&85.0&75.0&75.8&72.9&72.4&52.9&45.2&39.8&57.0&66.5&75.7&\textbf{61.0}&\textbf{59.3}\\
\hline
\end{tabular}}
\centering
\caption{Unseen-domain results (\%) of unsupervised single-domain (U), supervised lifelong (SL) and unsupervised lifelong (UL) methods. `BL' denotes our current domain baseline. `BL(GT)' refers to replacing pseudo labels with ground truth labels. As we do not have source code for re-implementation, GwFReID* is reported on Market$\to$Duke$\to$Cuhk-Sysu$\to$MSMT17, which benefit from more domain data than our setting.}
\label{table:unseen domains results}
\end{table}


\paragraph{Compared methods.}
We re-implement three types of methods for comparison on lifelong person ReID, including unsupervised single-domain methods, supervised lifelong methods and unsupervised lifelong methods. 

The unsupervised single-domain methods include SpCL~\cite{ge2020self} and ICE~\cite{Chen_2021_ICE}, which are trained sequentially on each seen domain. 

The lifelong methods include four general-purpose lifelong methods (LwF~\cite{Li2018LearningWF}, SPD~\cite{Tung2019SimilarityPreservingKD}, iCaRL~\cite{Rebuffi2017iCaRLIC} and C$o^2$L~\cite{cha2021co2l}) and two ReID-specific lifelong methods (AKA~\cite{pu_cvpr2021} and GwFReID~\cite{Wu2021GeneralisingWF}). LwF and SPD are pure distillation-based methods, which do not store old samples for rehearsal. LwF uses a prediction-level cross-entropy distillation~\cite{hinton2015distilling} between old and new domain models. SPD distills mid-level feature similarity between old and new domain models. iCaRL and C$o^2$L are rehearsal-based methods. iCaRL conducts prediction-level distillation on new and stored old images for rehearsal. C$o^2$L proposes an asymmetric supervised contrastive loss and a relation distillation for supervised continual learning.
For general-purpose methods LwF, SPD, iCaRL and C$o^2$L, we \textbf{combine the same current-domain contrastive baseline (Section~\ref{sec:Centroid-based Contrastive Baseline}) and the lifelong learning techniques of each paper} to convert these methods to person ReID and conduct a fair comparison with our method. For example, in LwF, we combine our contrastive baseline for learning current domain knowledge and the prediction-level distillation for mitigating the forgetting.

\begin{figure}[t]
\centering
   \includegraphics[width=1\linewidth]{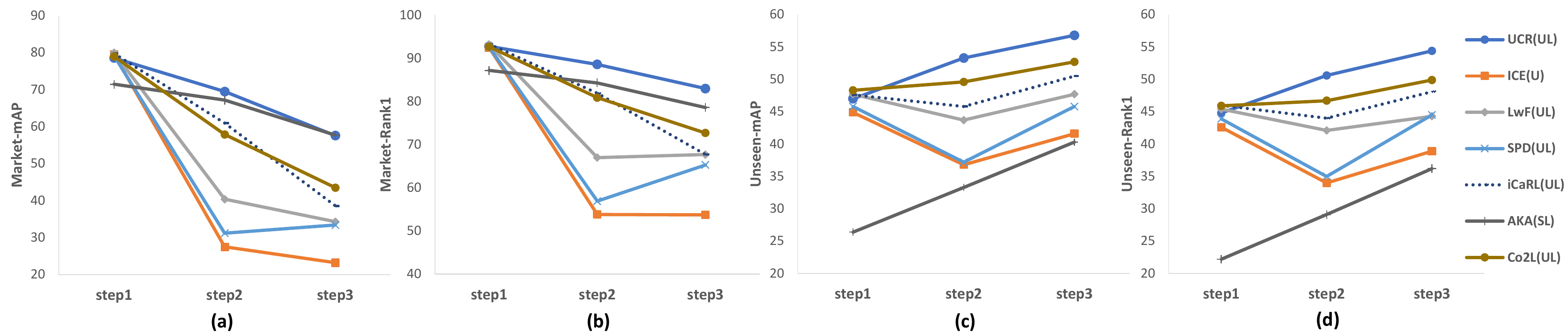}
   \caption{Non-forgetting evaluation on the first seen domain: (a) mAP and (b) Rank1 on Market-1501. Generalizability evaluation: (c) averaged mAP and (d) averaged Rank1 on the unseen domains.}
\label{fig:figure4}
\end{figure}

\paragraph{Seen-domain non-forgetting evaluation.}
We report seen-domain results after the final step in Table~\ref{table:seen domains results}. Designed for maximally learning domain-specific features inside a single domain, SpCL and ICE can not learn domain-agnostic generalized features for lifelong ReID. Among lifelong methods, the rehearsal-based methods iCaRL and C$o^2$L yield better averaged performance than the pure distillation-based methods LwF and SPD. Under the unsupervised lifelong setting, our proposed UCR outperforms the second best method C$o^2$L by 4.6\% on averaged mAP and 2.6\% on averaged Rank1. We also replace the pseudo labels with ground truth labels to compare in the supervised lifelong setting. Our re-implementation of LwF, SPD, iCaRL and C$o^2$L outperform the ReID-specific method AKA. Under the supervised lifelong setting, our proposed UCR outperform the second best method C$o^2$L by 3.5\% on averaged mAP and 2.4\% on averaged Rank1.  
We further draw first seen domain (Market) mAP/Rank1 variation curves after each step in Fig.~\ref{fig:figure4} (a) and (b), which confirm that our UCR has a slower forgetting rate.

\paragraph{Unseen-domain generalizability evaluation.}
To compare the generalizability of each method, we report unseen-domain results in Table~\ref{table:unseen domains results}. Similar to seen-domain results, SpCL and ICE can hardly learn domain-agnostic generalized features, which leads to low performance on unseen domains. On the contrary, lifelong methods accumulate knowledge from each domain and eventually learn domain-agnostic generalized features. 
With the same baseline, the rehearsal-based methods iCaRL and C$o^2$L outperform the pure distillation-based methods LwF and SPD. Under the unsupervised lifelong setting, our proposed UCR outperforms the second best method C$o^2$L by 4.1\% on averaged mAP and 4.5\% on averaged Rank1. We also compare the supervised performance of AKA, GwFReID, LwF, SPD, iCaRL and C$o^2$L on the unseen domains. Since the source code of GwFReID is not available, we report the results from the original paper, which benefit from one more seen domain (DukeMTMC) but are still lower than our results on iLIDS, CUHK03 and 3DPeS. Our proposed UCR outperforms the second best method C$o^2$L by 4.2\% on averaged mAP and 4.7\% on averaged Rank1. 
We draw averaged mAP/Rank1 variation curves on all the unseen domains after each step in Fig.~\ref{fig:figure4} (c) and (d), which show that our UCR achieves better generalizability than other methods.

\textbf{More analysis.} Our method shares certain similarity with C$o^2$L, because both methods use contrastive losses for rehearsal. However, C$o^2$L only uses current domain samples as anchors, while old domain samples are served as negatives in the contrastive loss. Our method UCR uses both current and old domain samples as anchors to retrieve the corresponding cluster prototypes from a prototype memory. In addition, we set a constraint to regularize the similarity relationship update, which is effective on similarity ranking problem person ReID. 

\subsection{Ablation study}
The performance improvement of UCR over the baseline mainly comes from our proposed old domain contrastive rehearsal and the image-to-image similarity constraint. To validate the effectiveness of each component, we conduct ablation experiments by gradually adding one of them to the baseline. In Table~\ref{table:ablation study}, `Baseline' refers to sequential training each new domain with $\mathcal{L}_{current}$ in Eq. (\ref{equ:current}) without any non-forgetting techniques. The performance on both seen and unseen domains can be improved by adding the old domain contrastive rehearsal `+$\mathcal{L}_{old}$'. The overall performance boost from the similarity constraint '+$\mathcal{L}_{sim}$' is more significant, which indicates that regularizing image-to-image relation is more effective than regularizing image-to-prototype relation for unsupervised lifelong ReID. One possible explanation is that prototypes are built by clustering pseudo labels, which can be noisy for lifelong ReID. `+$\mathcal{L}_{old}$+$\mathcal{L}_{sim}$' denotes our full UCR method. We conclude that the two components are always beneficial and complementary for both seen and unseen domains. 


\begin{table}[t]
\adjustbox{valign=t}{
\begin{minipage}[t]{0.35\textwidth} \centering
    \scalebox{0.7}{
    \begin{tabular}{c|cc|cc}
    \hline
\multirow{2}{*}{Method}  & \multicolumn{2}{c}{Seen-Avg} & \multicolumn{2}{|c}{Unseen-Avg} \\ \cline{2-5}
\multicolumn{1}{c|}{} & \multicolumn{1}{c}{mAP} & \multicolumn{1}{c|}{Rank1} & \multicolumn{1}{c}{mAP} & \multicolumn{1}{c}{Rank1} \\ \hline
Baseline&45.7&67.4&47.3&45.1\\
+$\mathcal{L}_{old}$&49.4&69.2&49.6&48.1\\
+$\mathcal{L}_{sim}$&53.7&73.6&54.7&51.3\\
+$\mathcal{L}_{old}$+$\mathcal{L}_{sim}$&\textbf{55.4}&\textbf{74.3}&\textbf{56.8}&\textbf{54.4}\\\hline
    \end{tabular}}
    \caption{Ablation study on the contrastive rehearsal loss $\mathcal{L}_{old}$ and the similarity constraint loss $\mathcal{L}_{sim}$. We report the averaged results.}
    \label{table:ablation study}
\end{minipage}}\hspace{0.05\textwidth}
\adjustbox{valign=t}{
\begin{minipage}[t]{0.6\textwidth} \centering
    \includegraphics[width=1\linewidth]{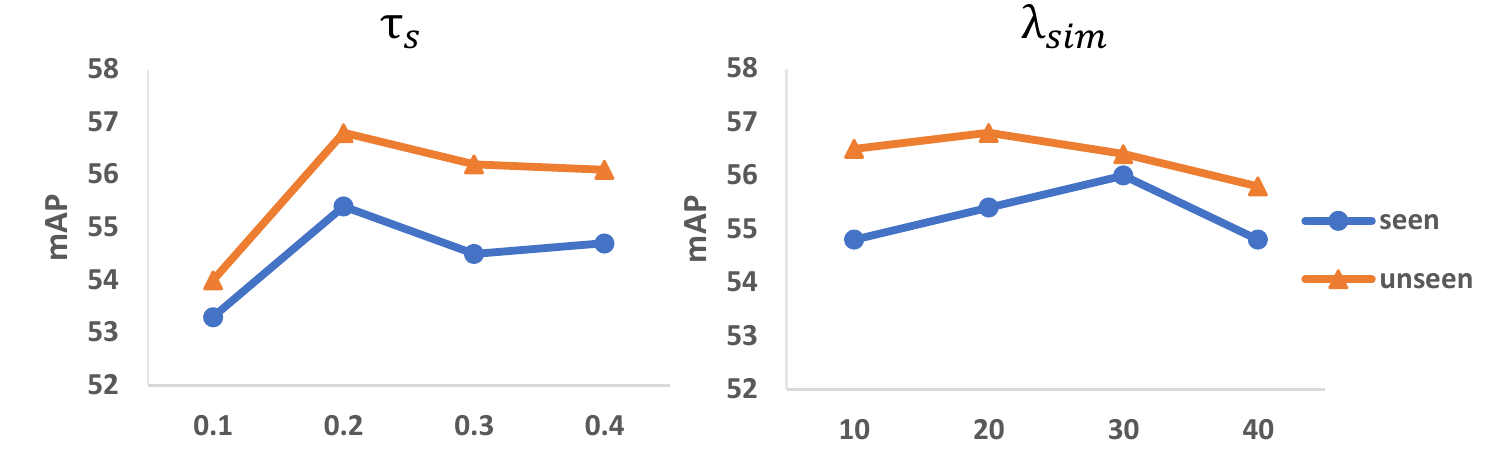}
    \captionof{figure}{Sensitivity to similarity constraint temperature $\tau_s$ and balancing coefficient $\lambda_{sim}$. We report the averaged results on seen and unseen domains.}
    \label{fig:figure3}
\end{minipage}}
\end{table}

\begin{table}[t]
\adjustbox{valign=t}{
\begin{minipage}[t]{0.3\textwidth} \centering
    \scalebox{0.7}{
    \begin{tabular}{c|cc|cc}
    \hline
    \multirow{2}{*}{$K_{mem}$}  & \multicolumn{2}{c}{Seen-Avg} & \multicolumn{2}{|c}{Unseen-Avg} \\ \cline{2-5}
    \multicolumn{1}{c|}{} & \multicolumn{1}{c}{mAP} & \multicolumn{1}{c|}{R1} & \multicolumn{1}{c}{mAP} & \multicolumn{1}{c}{R1} \\ \hline
    1&53.0&72.7&55.2&52.7\\
    2&55.4&74.3&56.8&54.4\\
    4&57.1&75.2&58.0&55.2\\
    8&58.1&75.8&\textbf{58.7}&55.6\\
    All&\textbf{59.3}&\textbf{75.9}&58.3&\textbf{55.9}\\
    \hline
    \end{tabular}}
    \caption{Number of images per pseudo identity in the memory.\vspace{-1em}}
    \label{table:K mem}
\end{minipage}}\hspace{0.02\textwidth}
\adjustbox{valign=t}{
\begin{minipage}[t]{0.3\textwidth} \centering
    \scalebox{0.7}{
    \begin{tabular}{c|cc|cc}
    \hline
    \multirow{2}{*}{$K_{mem}$}  & \multicolumn{2}{c}{Seen-Avg} & \multicolumn{2}{|c}{Unseen-Avg} \\ \cline{2-5}
    \multicolumn{1}{c|}{} & \multicolumn{1}{c}{mAP} & \multicolumn{1}{c|}{R1} & \multicolumn{1}{c}{mAP} & \multicolumn{1}{c}{R1} \\ \hline
    Nearest&\textbf{55.4}&\textbf{74.3}&\textbf{56.8}&\textbf{54.4}\\
    Farthest&55.1&74.2&55.5&53.4\\
    Random&55.0&73.7&55.3&53.6\\
    \hline
    \end{tabular}}
    \caption{memory sample selection based on the distance between images and the cluster prototype. }
    \label{table:sample selection}
\end{minipage}}\hspace{0.02\textwidth}
\adjustbox{valign=t}{
\begin{minipage}[t]{0.3\textwidth} \centering
    \scalebox{0.7}{
    \begin{tabular}{c|cc|cc}
    \hline
    \multirow{2}{*}{Method}  & \multicolumn{2}{c}{Seen-Avg} & \multicolumn{2}{|c}{Unseen-Avg} \\ \cline{2-5}
    \multicolumn{1}{c|}{} & \multicolumn{1}{c}{mAP} & \multicolumn{1}{c|}{R1} & \multicolumn{1}{c}{mAP} & \multicolumn{1}{c}{R1} \\ \hline
    C$o^2$L(UL)\cite{cha2021co2l}&53.7&66.1&50.5&46.7\\
    \textbf{UCR(UL)}&\textbf{55.3}&\textbf{71.6}&\textbf{52.1}&\textbf{49.6}\\\hline
    AKA(SL)~\cite{pu_cvpr2021}&43.9&58.9&41.1&38.1\\
    C$o^2$L(SL)\cite{cha2021co2l}&59.6&69.0&52.4&49.7\\
    \textbf{UCR(SL)}&\textbf{59.9}&\textbf{72.7}&\textbf{57.3}&\textbf{56.1}\\
    \hline
    \end{tabular}}
    \caption{Second training order: MSMT17 $\to$ Cuhk-Sysu $\to$ Market.}
    \label{table:order}
\end{minipage}}
\end{table}


\label{sec:Parameter analysis}
\subsection{Parameter analysis}
We analyze the sensitivity of our method to important hyper-parameters on lifelong person ReID performance. We first vary the number of stored images per identity $K_{mem}$ to study the sensitivity of our memory size. As shown in Table~\ref{table:K mem}, storing more samples with a larger $K_{mem}$ results in less forgetting and better generalizability. In our experiments, we set $K_{mem}=2$ to fairly compare with GwFReID~\cite{Wu2021GeneralisingWF}. In real-world deployments, users can choose a smaller $K_{mem}$ to save storage space and get slightly lower results, or a larger $K_{mem}$ to get better results. We further vary values of hyper-parameters $\tau_{s}$ and $\lambda_{sim}$ for the similarity constraint. Based on the results in Fig.~\ref{fig:figure3}, we set the similarity temperature $\tau_s=0.2$ and the balancing weight $\lambda_{sim}=20$. We analyze clustering parameters and compare backbone networks in Supplementary Materials.

\subsection{More discussion}

\paragraph{Data selection for memory update.} To update the image memory after each step, for each cluster, we can select $K_{mem}$ samples nearest to the cluster prototype, or farthest to the cluster prototype, or random samples. As shown in Table~\ref{table:sample selection}, storing nearest samples for rehearsal achieves slightly better performance, because nearest samples are more reliable clustered samples that bring in less pseudo label noise. 



\paragraph{Training order.} Our primary training order follows previous lifelong methods~\cite{Wu2021GeneralisingWF}, which starts from a medium domain Market and ends with the largest domain MSMT17. However, it is hard to control the order of upcoming domains in the real world. We test a second order MSMT17$\to$Cuhk-Sysu$\to$Market, which makes it easier to forget more knowledge on the largest domain MSMT17. As shown in Table~\ref{table:order}, UCR still significantly outperforms state-of-the-art methods AKA and C$o^2$L. Please refer to Supplementary Materials for more details.

\paragraph{Memory size.}
We use both image and prototype memory buffers in our rehearsal-based method UCR. With imperfect clustering pseudo labels and $K_{mem}=2$, our image memory stores approximately $640$(Market)$\times2+1050$(Cuhk-Sysu)$\times2+1900$(MSMT17)$\times2=7180$ images $\approx 11.8\%$ of all the training images (Market, Cuhk-Sysu and MSMT17). On the other hand, our prototype memory stores approximately $640$(Market)$+1050$(Cuhk-Sysu)$+1900$(MSMT17)$=3590$ prototype vectors (dimension $1\times2048\times1\times1$)$\approx$29.4 MB, which is negligible compared with storing dataset images (for example, MSMT17$\approx$2.5 GB). 

\paragraph{Limitation and future work.}
Even though we only store 2 images and 1 prototype representation vector per identity, the memory size can still grow rapidly due to daily recorded new data in real-world deployments. An interesting direction for our future work is to explore how to reduce the number of stored images. For example, setting up a metric to select more representative images and filter out less representative images from each domain can be a possible solution.



\section{Conclusion}
In this paper, we introduce a challenging but practical task, namely unsupervised lifelong person ReID, to explore the possibility of learning a generalizable model via sequential unsupervised adaptation on incrementally added domains. To tackle the catastrophic forgetting after the domain adaptation, we propose an unsupervised contrastive rehearsal (UCR) method, which rehearses a small number of old samples in a contrastive manner. We also set a similarity relationship constraint to regularize the model update in a way that suits old knowledge. In comparison with previous lifelong methods, our proposed UCR achieves better non-forgetting performance on seen domains and better generalizability on unseen domains. 




%
%
\bibliographystyle{splncs04}
\bibliography{egbib}

\begin{thebibliography}{10}
\providecommand{\url}[1]{\texttt{#1}}
\providecommand{\urlprefix}{URL }
\providecommand{\doi}[1]{https://doi.org/#1}

\bibitem{aljundi2018memory}
Aljundi, R., Babiloni, F., Elhoseiny, M., Rohrbach, M., Tuytelaars, T.: Memory
  aware synapses: Learning what (not) to forget. In: ECCV (2018)

\bibitem{3dpes}
Baltieri, D., Vezzani, R., Cucchiara, R.: 3dpes: 3d people dataset for
  surveillance and forensics. In: Proceedings of the 2011 joint ACM workshop on
  Human gesture and behavior understanding (2011)

\bibitem{Castro2018End}
Castro, F.M., Mar{\'i}n-Jim{\'e}nez, M.J., Guil, N., Schmid, C., Alahari, K.:
  End-to-end incremental learning. In: ECCV (2018)

\bibitem{cha2021co2l}
Cha, H., Lee, J., Shin, J.: Co2l: Contrastive continual learning. In: ICCV
  (2021)

\bibitem{chaudhry2018riemannian}
Chaudhry, A., Dokania, P.K., Ajanthan, T., Torr, P.H.: Riemannian walk for
  incremental learning: Understanding forgetting and intransigence. In: ECCV
  (2018)

\bibitem{Chen_2021_ICE}
Chen, H., Lagadec, B., Bremond, F.: Ice: Inter-instance contrastive encoding
  for unsupervised person re-identification. In: ICCV (2021)

\bibitem{Chen_2021_CVPR}
Chen, H., Wang, Y., Lagadec, B., Dantcheva, A., Bremond, F.: Joint generative
  and contrastive learning for unsupervised person re-identification. In: CVPR
  (2021)

\bibitem{chen2019abd}
Chen, T., Ding, S., Xie, J., Yuan, Y., Chen, W., Yang, Y., Ren, Z., Wang, Z.:
  Abd-net: Attentive but diverse person re-identification. In: ICCV (2019)

\bibitem{chen2020simple}
Chen, T., Kornblith, S., Norouzi, M., Hinton, G.: A simple framework for
  contrastive learning of visual representations. In: ICML (2020)

\bibitem{dai2021generalizable}
Dai, Y., Li, X., Liu, J., Tong, Z., Duan, L.Y.: Generalizable person
  re-identification with relevance-aware mixture of experts. In: CVPR (2021)

\bibitem{Dai_2021_ICCV}
Dai, Y., Liu, J., Sun, Y., Tong, Z., Zhang, C., Duan, L.Y.: Idm: An
  intermediate domain module for domain adaptive person re-id. In: ICCV (2021)

\bibitem{douillard2021plop}
Douillard, A., Chen, Y., Dapogny, A., Cord, M.: Plop: Learning without
  forgetting for continual semantic segmentation. In: CVPR (2021)

\bibitem{Ester1996ADA}
Ester, M., Kriegel, H.P., Sander, J., Xu, X.: A density-based algorithm for
  discovering clusters in large spatial databases with noise. In: KDD (1996)

\bibitem{ge2020mutual}
Ge, Y., Chen, D., Li, H.: Mutual mean-teaching: Pseudo label refinery for
  unsupervised domain adaptation on person re-identification. In: ICLR (2020)

\bibitem{ge2020self}
Ge, Y., Zhu, F., Chen, D., Zhao, R., Li, H.: Self-paced contrastive learning
  with hybrid memory for domain adaptive object re-id. In: NeurIPS (2020)

\bibitem{Gray2008ViewpointIP}
Gray, D., Tao, H.: Viewpoint invariant pedestrian recognition with an ensemble
  of localized features. In: ECCV (2008)

\bibitem{He_2020_CVPR}
He, K., Fan, H., Wu, Y., Xie, S., Girshick, R.: Momentum contrast for
  unsupervised visual representation learning. In: CVPR (2020)

\bibitem{he2016deep}
He, K., Zhang, X., Ren, S., Sun, J.: Deep residual learning for image
  recognition. In: CVPR (2016)

\bibitem{He_2021_ICCV}
He, S., Luo, H., Wang, P., Wang, F., Li, H., Jiang, W.: Transreid:
  Transformer-based object re-identification. In: ICCV (2021)

\bibitem{hinton2015distilling}
Hinton, G., Vinyals, O., Dean, J.: Distilling the knowledge in a neural
  network. arXiv preprint arXiv:1503.02531  (2015)

\bibitem{hirzer11}
Hirzer, M., Beleznai, C., Roth, P.M., Bischof, H.: {Person Re-Identification by
  Descriptive and Discriminative Classification}. In: {Proc. Scandinavian
  Conference on Image Analysis (SCIA)} (2011)

\bibitem{ioffe15batch}
Ioffe, S., Szegedy, C.: Batch normalization: Accelerating deep network training
  by reducing internal covariate shift. In: ICML (2015)

\bibitem{iscen2020memory}
Iscen, A., Zhang, J., Lazebnik, S., Schmid, C.: Memory-efficient incremental
  learning through feature adaptation. In: ECCV (2020)

\bibitem{Jin_2020_CVPR}
Jin, X., Lan, C., Zeng, W., Chen, Z., Zhang, L.: Style normalization and
  restitution for generalizable person re-identification. In: CVPR (2020)

\bibitem{Adam_optimization}
Kingma, D.P., Ba, J.: Adam: A method for stochastic optimization. In: ICLR
  (2015)

\bibitem{Li2013LocallyAF}
Li, W., Wang, X.: Locally aligned feature transforms across views. CVPR  (2013)

\bibitem{Li2012HumanRW}
Li, W., Zhao, R., Wang, X.: Human reidentification with transferred metric
  learning. In: ACCV (2012)

\bibitem{Li2014DeepReIDDF}
Li, W., Zhao, R., Xiao, T., Wang, X.: Deepreid: Deep filter pairing neural
  network for person re-identification. CVPR  (2014)

\bibitem{Li2018LearningWF}
Li, Z., Hoiem, D.: Learning without forgetting. IEEE TPAMI  (2018)

\bibitem{Loy2009MulticameraAC}
Loy, C.C., Xiang, T., Gong, S.: Multi-camera activity correlation analysis. In:
  CVPR (2009)

\bibitem{Luo_2019_CVPR_Workshops}
Luo, H., Gu, Y., Liao, X., Lai, S., Jiang, W.: Bag of tricks and a strong
  baseline for deep person re-identification. In: CVPR Workshops (2019)

\bibitem{vandermaaten08a}
van~der Maaten, L., Hinton, G.: Visualizing data using t-sne. JMLR  (2008)

\bibitem{pan2018two}
Pan, X., Luo, P., Shi, J., Tang, X.: Two at once: Enhancing learning and
  generalization capacities via ibn-net. In: ECCV (2018)

\bibitem{PyTorch_NEURIPS2019}
Paszke, A., Gross, S., Massa, F., Lerer, A., Bradbury, J., Chanan, G., Killeen,
  T., Lin, Z., Gimelshein, N., Antiga, L., Desmaison, A., Kopf, A., Yang, E.,
  DeVito, Z., Raison, M., Tejani, A., Chilamkurthy, S., Steiner, B., Fang, L.,
  Bai, J., Chintala, S.: Pytorch: An imperative style, high-performance deep
  learning library. In: NeurIPS (2019)

\bibitem{pu_cvpr2021}
Pu, N., Chen, W., Liu, Y., Bakker, E.M., Lew, M.S.: Lifelong person
  re-identification via adaptive knowledge accumulation. In: CVPR (2021)

\bibitem{Rebuffi2017iCaRLIC}
Rebuffi, S.A., Kolesnikov, A., Sperl, G., Lampert, C.H.: icarl: Incremental
  classifier and representation learning. CVPR  (2017)

\bibitem{ristani2016MTMC}
Ristani, E., Solera, F., Zou, R., Cucchiara, R., Tomasi, C.: Performance
  measures and a data set for multi-target, multi-camera tracking. In: ECCV
  workshops (2016)

\bibitem{rostami2021lifelong}
Rostami, M.: Lifelong domain adaptation via consolidated internal distribution.
  NeurIPS  (2021)

\bibitem{Russakovsky2015ImageNetLS}
Russakovsky, O., Deng, J., Su, H., Krause, J., Satheesh, S., Ma, S., Huang, Z.,
  Karpathy, A., Khosla, A., Bernstein, M., Berg, A., Fei-Fei, L.: Imagenet
  large scale visual recognition challenge. IJCV  (2015)

\bibitem{Song2019GeneralizablePR}
Song, J., Yang, Y., Song, Y.Z., Xiang, T., Hospedales, T.M.: Generalizable
  person re-identification by domain-invariant mapping network. CVPR  (2019)

\bibitem{tang2021gradient}
Tang, S., Su, P., Chen, D., Ouyang, W.: Gradient regularized contrastive
  learning for continual domain adaptation. In: AAAI (2021)

\bibitem{Tung2019SimilarityPreservingKD}
Tung, F., Mori, G.: Similarity-preserving knowledge distillation. ICCV  (2019)

\bibitem{Wang2021camawareproxies}
Wang, M., Lai, B., Huang, J., Gong, X., Hua, X.S.: Camera-aware proxies for
  unsupervised person re-identification. In: AAAI (2021)

\bibitem{wei2018person}
Wei, L., Zhang, S., Gao, W., Tian, Q.: Person transfer gan to bridge domain gap
  for person re-identification. In: CVPR (2018)

\bibitem{Wu2021GeneralisingWF}
Wu, G., Gong, S.: Generalising without forgetting for lifelong person
  re-identification. In: AAAI (2021)

\bibitem{Wu2018UnsupervisedFL}
Wu, Z., Xiong, Y., Yu, S.X., Lin, D.: Unsupervised feature learning via
  non-parametric instance discrimination. In: CVPR (2018)

\bibitem{Xiao2017JointDA}
Xiao, T., Li, S., Wang, B., Lin, L., Wang, X.: Joint detection and
  identification feature learning for person search. CVPR  (2017)

\bibitem{yoon2018lifelong}
Yoon, J., Yang, E., Lee, J., Hwang, S.J.: Lifelong learning with dynamically
  expandable networks. In: ICLR (2018)

\bibitem{Zhao2017SpindleNP}
Zhao, H., Tian, M., Sun, S., Shao, J., Yan, J., Yi, S., Wang, X., Tang, X.:
  Spindle net: Person re-identification with human body region guided feature
  decomposition and fusion. CVPR  (2017)

\bibitem{Zheng2015ScalablePR}
Zheng, L., Shen, L., Tian, L., Wang, S., Wang, J., Tian, Q.: Scalable person
  re-identification: A benchmark. ICCV  (2015)

\bibitem{Zheng2009AssociatingGO}
Zheng, W.S., Gong, S., Xiang, T.: Associating groups of people. In: BMVC (2009)

\bibitem{zheng2019joint}
Zheng, Z., Yang, X., Yu, Z., Zheng, L., Yang, Y., Kautz, J.: Joint
  discriminative and generative learning for person re-identification. In: CVPR
  (2019)

\bibitem{zhong2017re}
Zhong, Z., Zheng, L., Cao, D., Li, S.: Re-ranking person re-identification with
  k-reciprocal encoding. In: CVPR (2017)

\bibitem{Zhong2020RandomED}
Zhong, Z., Zheng, L., Kang, G., Li, S., Yang, Y.: Random erasing data
  augmentation. In: AAAI (2020)

\end{thebibliography}

\appendix

\chapter*{Supplementary Materials}

In the supplementary material, we provide more details about our algorithm in Section~\ref{Algorithm Details}, the possible alternatives for our current domain contrastive baseline in Section~\ref{Current domain contrastive baseline} and the performance of second training order in Section~\ref{Second training order}. We proceed to analyze the generalizability of our method after each training step with a domain gap visualization in Section~\ref{Domain gap visualization}. We further analyze the sensitivity to clustering hyper-parameters in Section~\ref{Clustering distance threshold} and the performance with a stronger backbone network in Section~\ref{Backbone Network}.

\section{Algorithm Details}
\label{Algorithm Details}
Our proposed UCR is composed of one fully unsupervised learning step on the first domain $D_1$ and several unsupervised incremental learning steps on the following domains $D_2, ..., D_N$. In this way, we incrementally accumulate knowledge from each seen domain into a same momentum encoder $\theta_m$. To help readers better understand our proposed method, we present algorithm details in Algorithm~\ref{algo:1}. 

\begin{algorithm}
    \SetKwInOut{Input}{Input}
    \Input{Unlabeled seen domains $D_1, D_2, ..., D_N$.  }
    \For{$domain=D_1$ to $D_N$}{
    \eIf{First domain $D_1$}
    {
        \textbf{\# Single-domain unsupervised learning \#} \\
        \For{$epoch=1$ to $E_{max}$}
      {
        Generate pseudo labels on $D_1$\;
        Calculate current domain prototypes $P^{c}$ in Eq.~(\textcolor{black}{3}) and camera prototypes in Eq.~(\textcolor{black}{5}) on $D_1$\;
        Initialize prototype memory $P=P^{c}$\;
        \For{$iter=1$ to $I_{max}$}
        {
         Sample a mini-batch from current domain $D_1$\ for $\mathcal{L}_{current}$ in Eq.~(\textcolor{black}{7})\;
         Update $\theta$ with $\mathcal{L}_{current}$\;
         Update $\theta_m$ with Eq.~(\textcolor{black}{1})\;
        }
      }
      Initialize image memory with $K_{mem}$ images per identity\;
    }
    {
        \textbf{\# Unsupervised lifelong learning with rehearsal \#} \\
        \For{$epoch=1$ to $E_{max}$}
      {
        Generate pseudo labels on $D_i$\;
        Calculate current domain prototypes $P^{c}$ in Eq.~(\textcolor{black}{3}) and camera prototypes in Eq.~(\textcolor{black}{5}) on $D_i$\;
        Update prototype memory $P=P^{o}\cup P^{c}$\;
        \For{$iter=1$ to $I_{max}$}
        {
         Sample a mini-batch from current domain $D_i$\ for $\mathcal{L}_{current}$ in Eq.~(\textcolor{black}{7})\;
         Sample a mini-batch from image memory for $\mathcal{L}_{old}$ in Eq.~(\textcolor{black}{8}) and $\mathcal{L}_{sim}$ in Eq.~(\textcolor{black}{11})\;
         Train $\theta$ with $\mathcal{L}_{overall}$ in Eq.~(\textcolor{black}{2})\;
         Update $\theta_m$ with Eq.~(\textcolor{black}{1})\;
        }
      }
      Update image memory with $K_{mem}$ images per identity\;
    }
    $\theta_f\gets \theta_m$ and $\theta \gets \theta_m$\;
    }
    \Return $\theta_m$ after the final domain $D_N$
    \caption{UCR for unsupervised lifelong person ReID. }
    \label{algo:1}
\end{algorithm}

\setcounter{equation}{11}
\setcounter{table}{7}
\setcounter{figure}{4}

\section{Current domain contrastive baseline}
\label{Current domain contrastive baseline}
For unsupervised lifelong person ReID, purifying the intra-cluster variance in each current domain helps to mitigate the noise in the knowledge accumulation.

In our main paper, the current domain baseline is defined as $\mathcal{L}_{current} = \mathcal{L}_{cluster}+\lambda_{cam} \mathcal{L}_{cam}$, where $\mathcal{L}_{cam}$ uses camera labels to purify the camera style variance in current domain knowledge. Actually, $\mathcal{L}_{cam}$ can be replaced by other variance reduction techniques, such as hard instance contrastive loss~\cite{Chen_2021_ICE}. The hard instance contrastive loss $\mathcal{L}_{hard}$ maximizes the similarity between an anchor representation $f(x_{i}^{c}|\theta)$ (superscript c denotes the current domain) and the hardest pseudo-positive $f(x_{hard}^{c}|\theta_m)$ in the mini-batch: 
\begin{equation}
   \mathcal{L}_{hard} = \mathop{\mathbb{E}}[-\log{\frac{\exp{(<f(x_{i}^{c}|\theta) \cdot f(x_{hard}^{c}|\theta_m)>)}}{\sum\nolimits_{j=1}^{N_{neg}+1}\exp{(<f(x_{i}^{c}|\theta) \cdot f(x_{j}^{c}|\theta_m)>)}}}]
\label{equa:viewloss}
\end{equation}
where $< \cdot >$ denotes the normalized cosine similarity. $f(x_{hard}^{c}|\theta_m)$ is the pseudo-positive that has the lowest cosine similarity with $f(x_{i}^{c}|\theta)$. $N_{neg}$ is the number of pseudo-negatives in the mini-batch. $\mathcal{L}_{hard}$ reduces the distance between anchor and hard samples to encourage the compactness of a cluster, so that intra-cluster variance can be reduced in the current domain. 

We compare three possible baselines for learning current domain knowledge: 1) only contrasting general cluster prototypes $\mathcal{L}_{current} = \mathcal{L}_{cluster}$, 2) reducing current domain noise by mining hard positives $\mathcal{L}_{current} = \mathcal{L}_{cluster}+\mathcal{L}_{hard}$ and 3) reducing current domain noise with camera labels $\mathcal{L}_{current} = \mathcal{L}_{cluster}+\mathcal{L}_{cam}$. As shown in Table~\ref{table:ablation study with 3 baselines}, the cluster prototypes in the case 1 can be noisy if we do not reduce intra-cluster variance, leading to unsatisfactory rehearsal effectiveness of $\mathcal{L}_{old}$. Reducing the distance between hard samples inside a cluster with $\mathcal{L}_{hard}$ (case 2) can mitigate the noise in cluster prototypes. The most effective way is to leverage camera labels with $\mathcal{L}_{cam}$ (case 3), as presented in the main paper.

\begin{table}
\scalebox{0.9}{
\begin{tabular}{c|c|cc|cc}
\hline
\multirow{2}{*}{Baseline} & \multirow{2}{*}{Method}  & \multicolumn{2}{c}{Seen-Avg} & \multicolumn{2}{|c}{Unseen-Avg} \\ \cline{3-6}
\multicolumn{1}{c|}{} &\multicolumn{1}{c|}{} & \multicolumn{1}{c}{mAP} & \multicolumn{1}{c|}{Rank1} & \multicolumn{1}{c}{mAP} & \multicolumn{1}{c}{Rank1} \\ 
\hline
\multirow{4}{*}{1) $\mathcal{L}_{current} = \mathcal{L}_{cluster}$}&$\mathcal{L}_{current}$&37.9&55.7&38.8&37.3\\
&+$\mathcal{L}_{old}$&38.7&54.7&34.7&33.5\\
&+$\mathcal{L}_{sim}$&43.9&62.6&44.0&42.8\\
&+$\mathcal{L}_{old}$+$\mathcal{L}_{sim}$&44.9&62.3&42.9&42.0\\
\hline
\multirow{4}{*}{2) $\mathcal{L}_{current} = \mathcal{L}_{cluster}+\mathcal{L}_{hard}$}&$\mathcal{L}_{current}$&38.8&58.0&44.5&41.9\\
&+$\mathcal{L}_{old}$&47.0&65.2&48.4&46.5\\
&+$\mathcal{L}_{sim}$&53.0&67.2&54.3&50.7\\
&+$\mathcal{L}_{old}$+$\mathcal{L}_{sim}$&52.6&68.7&55.1&52.7\\
\hline
\multirow{4}{*}{3) $\mathcal{L}_{current} = \mathcal{L}_{cluster}+\mathcal{L}_{cam}$}&$\mathcal{L}_{current}$&45.7&67.4&47.3&45.1\\
\multirow{4}{*}{(used in main paper)}&+$\mathcal{L}_{old}$&49.4&69.2&49.6&48.1\\
&+$\mathcal{L}_{sim}$&53.7&73.6&54.7&51.3\\
&+$\mathcal{L}_{old}$+$\mathcal{L}_{sim}$&\textbf{55.4}&\textbf{74.3}&\textbf{56.8}&\textbf{54.4}\\
\hline

\end{tabular}}
\centering
\caption{Ablation study on the old domain contrastive rehearsal loss $\mathcal{L}_{old}$ and the similarity distillation loss $\mathcal{L}_{sim}$ on the alternative baselines. We report the averaged results on seen and unseen domains.}
\label{table:ablation study with 3 baselines}
\end{table}

\section{Second training order}
\label{Second training order}
To complement Table 7 in the main paper, we provide more details about the performance on each dataset under the second training order MSMT17$\to$Cuhk-Sysu$\to$Market. The second training order starts from the largest dataset MSMT17 and ends by a medium dataset Market, which is opposite to our primary training order Market$\to$Cuhk-Sysu$\to$MSMT17. As shown in Table~\ref{table:seen domains results second order} and Table~\ref{table:unseen domains results second order}, our method outperforms state-of-the-art methods AKA and C$o^2$L on both seen and unseen domains by a clear margin. Our UCR(UL) yields better non-forgetting performance on the first domain MSMT17 than UCR(SL), because the clustering in UCR(UL) generates approximately 1050 pseudo-identitites for Cuhk-Sysu, while UCR(SL) contains 5532 ground truth identitites for Cuhk-Sysu. UCR(SL) stores more cluster prototypes and images for Cuhk-Sysu, which decreases the weight of MSMT17 in the memory buffers. 

\begin{table}[t]
\scalebox{0.8}{
\begin{tabular}{l|c|c|cc|cc|cc|cc}
\hline
\multicolumn{11}{c}{Second training order: MSMT17$\to$Cuhk-Sysu$\to$Market}\\\hline
\multirow{2}{*}{Method} & Memory & \multirow{2}{*}{Type} & \multicolumn{2}{c}{MSMT17} & \multicolumn{2}{|c}{Cuhk-Sysu} & \multicolumn{2}{|c}{Market} & \multicolumn{2}{|c}{Average} \\ \cline{4-11}
\multicolumn{1}{c}{} &\multicolumn{1}{|c|}{(image per id)}&\multicolumn{1}{c|}{}& \multicolumn{1}{c}{mAP} & \multicolumn{1}{c|}{Rank1} &\multicolumn{1}{c}{mAP} & \multicolumn{1}{c|}{Rank1} &\multicolumn{1}{c}{mAP} & \multicolumn{1}{c|}{Rank1} &\multicolumn{1}{c}{mAP} & \multicolumn{1}{c}{Rank1}  \\ 
\hline
BL+C$o^2$L~\cite{cha2021co2l}&2&UL&9.6&27.2&77.3&79.6&74.4&91.4&53.7&66.1\\
BL+\textbf{UCR}&2&UL&15.7&41.3&81.1&83.7&69.2&89.9&\textbf{55.3}&\textbf{71.6}\\
\hline
AKA~\cite{pu_cvpr2021}&0&SL&13.4&31.6&74.5&77.9&43.8&67.1&43.9&58.9\\
BL(GT)+C$o^2$L~\cite{cha2021co2l}&2&SL&9.9&26.4&83.9&85.7&84.9&94.8&59.6&69.0\\
BL(GT)+\textbf{UCR}&2&SL&12.7&36.3&87.9&89.5&79.0&92.2&\textbf{59.9}&\textbf{72.7}\\
\hline
\end{tabular}}
\centering
\caption{Results (\%) of supervised lifelong methods (SL) and unsupervised lifelong methods (UL) on seen domains. `BL' denotes our current domain baseline. `BL(GT)' refers to replacing pseudo labels with ground truth labels.}
\label{table:seen domains results second order}
\end{table}

\begin{table}[t]
\scalebox{0.6}{
\begin{tabular}{l|c|c|cc|cc|cc|cc|cc|cc|cc|cc|cc|cc}
\hline
\multicolumn{23}{c}{Second training order: MSMT17$\to$Cuhk-Sysu$\to$Market}\\\hline
\multirow{2}{*}{Method} & Memory&\multirow{2}{*}{Type}& \multicolumn{2}{|c}{VIPeR} & \multicolumn{2}{|c}{PRID} & \multicolumn{2}{|c}{GRID} & \multicolumn{2}{|c}{iLIDS} & \multicolumn{2}{|c}{CUHK01}& \multicolumn{2}{|c}{CUHK02}& \multicolumn{2}{|c}{SenseReID}& \multicolumn{2}{|c}{CUHK03}& \multicolumn{2}{|c}{3DPeS}& \multicolumn{2}{|c}{Average} \\ \cline{4-22}
\multicolumn{1}{c}{}&\multicolumn{1}{|c|}{(per id)}&\multicolumn{1}{|c}{}& \multicolumn{1}{|c}{mAP} & \multicolumn{1}{c|}{R1} &\multicolumn{1}{c}{mAP} & \multicolumn{1}{c|}{R1} &\multicolumn{1}{c}{mAP} & \multicolumn{1}{c|}{R1} &\multicolumn{1}{c}{mAP} & \multicolumn{1}{c|}{R1} &\multicolumn{1}{c}{mAP} & \multicolumn{1}{c|}{R1} &\multicolumn{1}{c}{mAP} & \multicolumn{1}{c|}{R1} &\multicolumn{1}{c}{mAP} & \multicolumn{1}{c|}{R1} &\multicolumn{1}{c}{mAP} & \multicolumn{1}{c|}{R1} &\multicolumn{1}{c}{mAP} & \multicolumn{1}{c|}{R1} &\multicolumn{1}{c}{mAP} & \multicolumn{1}{c}{R1}  \\ 
\hline
BL+C$o^2$L~\cite{cha2021co2l}&2&UL&43.7&33.5&43.0&32.0&40.0&31.2&76.9&68.3&61.9&61.9&61.2&57.1&43.3&35.7&26.7&36.7&57.5&63.4&50.5&46.7\\
BL+\textbf{UCR}&2&UL&43.4&34.8&50.7&40.0&32.2&24.0&80.6&73.3&66.5&66.5&62.8&60.3&45.8&38.9&26.9&41.2&60.3&67.3&\textbf{52.1}&\textbf{49.6}\\
\hline
AKA~\cite{pu_cvpr2021}&0&SL&36.3&26.3&38.0&29.0&18.7&13.6&69.6&60.0&60.9&61.7&53.9&54.2&30.2&24.6&19.2&19.5&43.1&54.0&41.1&38.1\\
BL(GT)+C$o^2$L~\cite{cha2021co2l}&2&SL&46.6&37.7&44.5&34.0&39.6&30.4&77.2&68.3&64.6&63.6&64.5&63.0&48.4&40.3&30.1&44.7&55.9&65.8&52.4&49.7\\
BL(GT)+\textbf{UCR}&2&SL&51.3&41.8&50.3&37.0&43.1&33.6&85.6&80.0&73.9&73.4&73.2&73.6&49.5&41.2&28.7&53.2&59.8&70.8&\textbf{57.3}&\textbf{56.1}\\
\hline
\end{tabular}}
\centering
\caption{Results (\%) of supervised lifelong methods (SL) and unsupervised lifelong methods (UL) on unseen domains. `BL' denotes our current domain baseline. `BL(GT)' refers to replacing pseudo labels with ground truth labels.}
\label{table:unseen domains results second order}
\end{table}

\section{Domain gap visualization}
\label{Domain gap visualization}
\begin{figure}
\centering
   \includegraphics[width=1\linewidth]{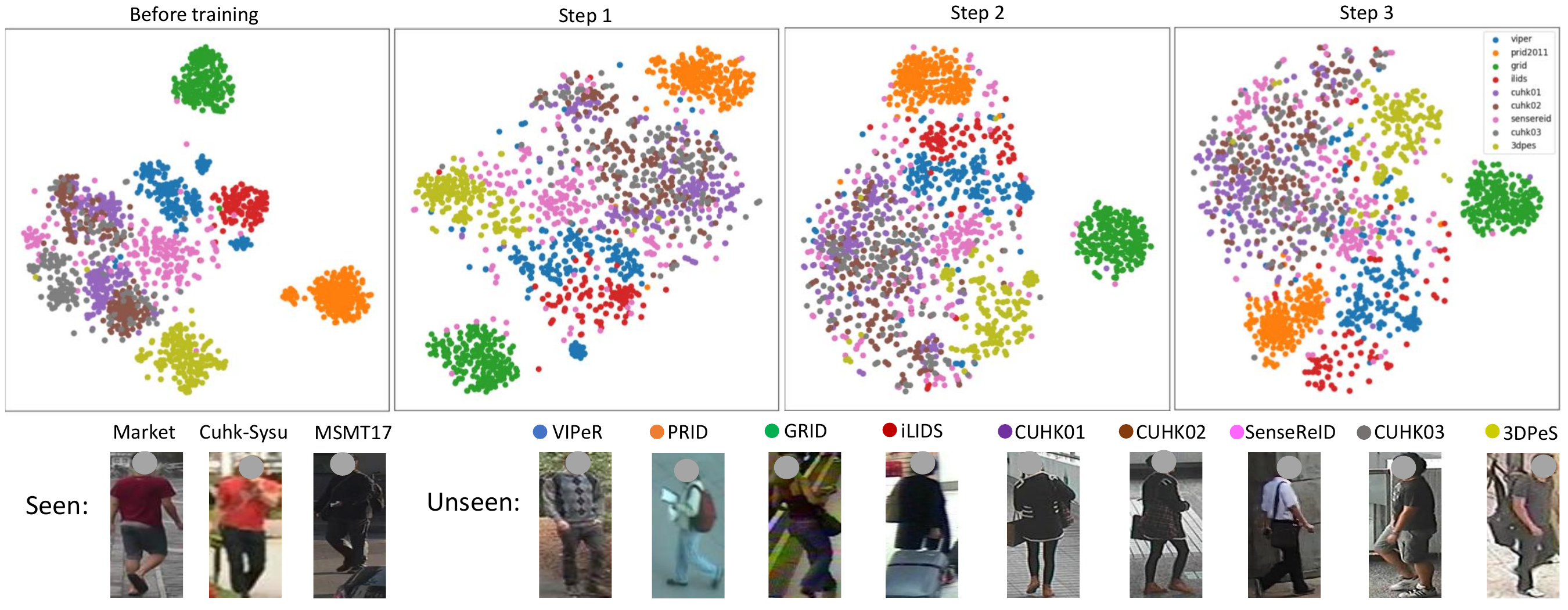}
   \caption{T-SNE visualization of unseen-domain representations after each step.}
\label{fig:figure6}
\end{figure}

We use t-SNE~\cite{vandermaaten08a} visualization on 200 randomly selected samples from each unseen domain to roughly estimate the domain gap encoded in representations after each training step. As shown in Figure~\ref{fig:figure6}, the domain gap is obvious before training, especially on GRID (\textcolor{ao(english)}{green}), iLIDS (\textcolor{red}{red}), PRID (\textcolor{orange}{orange}) and 3DPeS (\textcolor{olive}{olive}). The first step transfers our model from ImageNet distribution into Market person ReID distribution, which generally reduces the domain gap. The second step accumulates more domain knowledge into our model, making PRID (\textcolor{orange}{orange}) and 3DPeS (\textcolor{olive}{olive}) get closer to other domains. As MSMT17 contains more illumination and scenario diversity, the third step further reduces the domain gap, for instance, between GRID (\textcolor{ao(english)}{green}) and other domains. GRID (\textcolor{ao(english)}{green}) is recorded in underground stations that have an illumination level and backgrounds significantly different to other street camera recorded datasets. We conclude that our unsupervised lifelong method UCR effectively reduces the domain gap encoded in representations and incrementally learns domain-agnostic features. 

\section{Clustering distance threshold}
\label{Clustering distance threshold}
The distance threshold of DBSCAN is the maximal distance between two samples for one to be considered as in the neighborhood of the other. A larger distance threshold enlarges the radius of a cluster, making more samples be considered into a same cluster.

Our proposed method UCR conducts image-to-prototype contrastive learning with clustering generated pseudo labels, which can be affected by the clustering distance threshold. In our main paper, we follow ICE~\cite{Chen_2021_ICE} to set the threshold to 0.55. As it is hard to decide a uniform threshold for upcoming datasets of different sizes, we evaluate the sensitivity of UCR to the clustering distance threshold. We can observe from Table~\ref{table:distance threshold} that different threshold values only bring slight seen-domain and unseen-domain performance variation, showing that UCR is robust to the clustering threshold.

\begin{table}
\scalebox{0.9}{
\begin{tabular}{c|cc|cc}
\hline
\multirow{2}{*}{Threshold}  & \multicolumn{2}{c}{Seen-Avg} & \multicolumn{2}{|c}{Unseen-Avg} \\ \cline{2-5}
\multicolumn{1}{c|}{} & \multicolumn{1}{c}{mAP} & \multicolumn{1}{c|}{R1} & \multicolumn{1}{c}{mAP} & \multicolumn{1}{c}{R1} \\ \hline
0.45&55.0&73.4&56.6&54.3\\
0.5&55.3&74.0&\textbf{57.2}&\textbf{54.7}\\
0.55&55.4&74.3&56.8&54.4\\
0.6&\textbf{56.4}&\textbf{74.9}&56.4&54.1\\
0.65&55.4&73.4&55.6&52.1\\
\hline
\end{tabular}}
\centering
\caption{DBSCAN clustering distance threshold.}
\label{table:distance threshold}
\end{table}

\section{Backbone Network}
\label{Backbone Network}
Our proposed unsupervised lifelong method UCR leverages multiple domains to learn generalized features that can achieve balanced performance on all the domains. The generalizability is strongly related to the backbone network. A ResNet50 backbone is used in the experiments of our main paper to have a fair comparison with previous methods. In fact, the performance of UCR can be further improved with a backbone with stronger generalizability, such as IBN-ResNet50~\cite{pan2018two}. In IBN-ResNet50, authors propose to replace batch normalization~\cite{ioffe15batch} in ResNet with instance-batch normalization to enhance model generalizability. We compare the performance of using ResNet50 and IBN-ResNet50 as our backbone network in Table~\ref{table:backbone network}. IBN-ResNet50 significantly fills the performance gap between unsupervised and supervised lifelong settings. Under unsupervised lifelong setting, IBN-ResNet50 outperforms ResNet50 by a large margin. Under supervised lifelong setting, IBN-ResNet50 outperforms ResNet50 on unseen domains but not on seen domains, which indicates that instance-batch normalization brings in better generalizability rather than non-forgetting capacity when there is no label noise. 

\begin{table}
\scalebox{0.9}{
\begin{tabular}{c|c|cc|cc}
\hline
\multirow{2}{*}{Backbone}  &\multirow{2}{*}{Type} & \multicolumn{2}{c}{Seen-Avg} & \multicolumn{2}{|c}{Unseen-Avg} \\ \cline{3-6}
\multicolumn{1}{c|}{} &\multicolumn{1}{c|}{} & \multicolumn{1}{c}{mAP} & \multicolumn{1}{c|}{R1} & \multicolumn{1}{c}{mAP} & \multicolumn{1}{c}{R1} \\ \hline
ResNet50&UL&55.4&74.3&56.8&54.4\\
IBN-ResNet50&UL&\textbf{59.4}&\textbf{77.5}&\textbf{61.1}&\textbf{58.5}\\
\hline
ResNet50+GT&SL&\textbf{62.8}&80.1&61.0&59.3\\
IBN-ResNet50+GT&SL&61.9&\textbf{80.2}&\textbf{62.7}&\textbf{60.9}\\
\hline
\end{tabular}}
\centering
\caption{Comparison of backbone networks in our proposed UCR under unsupervised lifelong (UL) and supervised lifelong (SL) settings. `GT' refers to ground truth.}
\label{table:backbone network}
\end{table}


\end{document}